\documentclass[10pt,twocolumn,letterpaper]{article}

\usepackage[pagenumbers]{cvpr} 

\definecolor{cvprblue}{rgb}{0.21,0.49,0.74}
\usepackage[pagebackref,breaklinks,colorlinks,allcolors=cvprblue]{hyperref}
\usepackage{hypernat}
\usepackage{booktabs}
\usepackage{multirow}
\usepackage{graphicx}
\usepackage[table]{xcolor}
\usepackage{arydshln}
\usepackage{cuted}
\usepackage{caption}
\usepackage{subcaption}
\usepackage{makecell}

\usepackage{amsmath,amssymb}
\usepackage{algorithm}
\usepackage{algpseudocode}
\usepackage{cuted}

\def\confName{CVPR}
\def\confYear{2027}

\title{FastLayered: Training-Free Acceleration for Image-to-Layer Generation via Composite-Guided Velocity Calibration}

\author{
Changwang Mei\textsuperscript{1}
\qquad
Peisong Wang\textsuperscript{2,$\dagger$}
\qquad
Meng Li\textsuperscript{1}
\qquad
Zekun Li\textsuperscript{2,3}
\\
Zehua Wang\textsuperscript{1}
\qquad
Zhihui Wei\textsuperscript{1,$\dagger$}
\qquad
Jian Cheng\textsuperscript{2,$\dagger$}
\\
\textsuperscript{1}Nanjing University of Science and Technology, Nanjing, China
\\
\textsuperscript{2}C\textsuperscript{2}DL,
Institute of Automation, Chinese Academy of Sciences, Beijing, China
\\
\textsuperscript{3}Beijing Academy of Artificial Intelligence, Beijing, China
\\[2pt]
{\ttfamily\small
\{meichangwang,limeng2024,gswei\}@njust.edu.cn
}
\\
{\ttfamily\small
relaxwang0714@gmail.com
\qquad
\{peisong.wang,jcheng\}@nlpr.ia.ac.cn
}
}
\begin{document}
\maketitle

\begingroup
\renewcommand{\thefootnote}{\fnsymbol{footnote}}
\footnotetext[2]{Corresponding authors.}
\endgroup

\begin{strip}
    \centering
    \includegraphics[width=\linewidth]{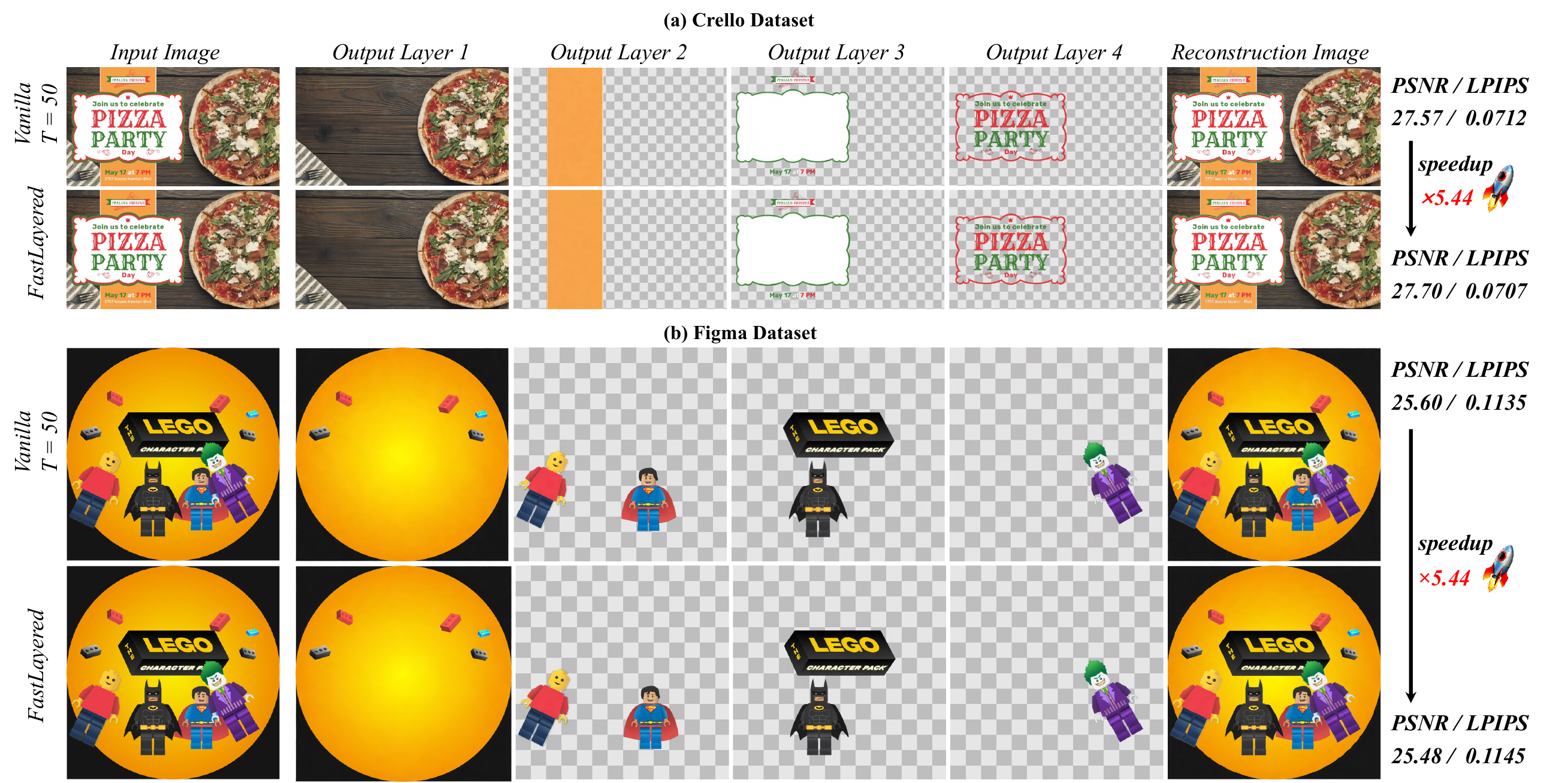}

    \captionsetup{hypcap=false}
    \captionof{figure}{
        Qualitative comparison between FastLayered and vanilla inference with 50 sampling steps on the Crello and Figma datasets using Qwen-Image-Layered. FastLayered maintains comparable layer decomposition quality and global reconstruction fidelity.
    }
    \captionsetup{hypcap=true}
    \label{fig:teaser}
\end{strip}

\begin{figure*}[t]
    \centering

    \begin{minipage}[c]{0.48\textwidth}
        \centering

        \begin{subfigure}[b]{0.7\linewidth}
            \centering
            \includegraphics[
                width=\linewidth,
                keepaspectratio
            ]{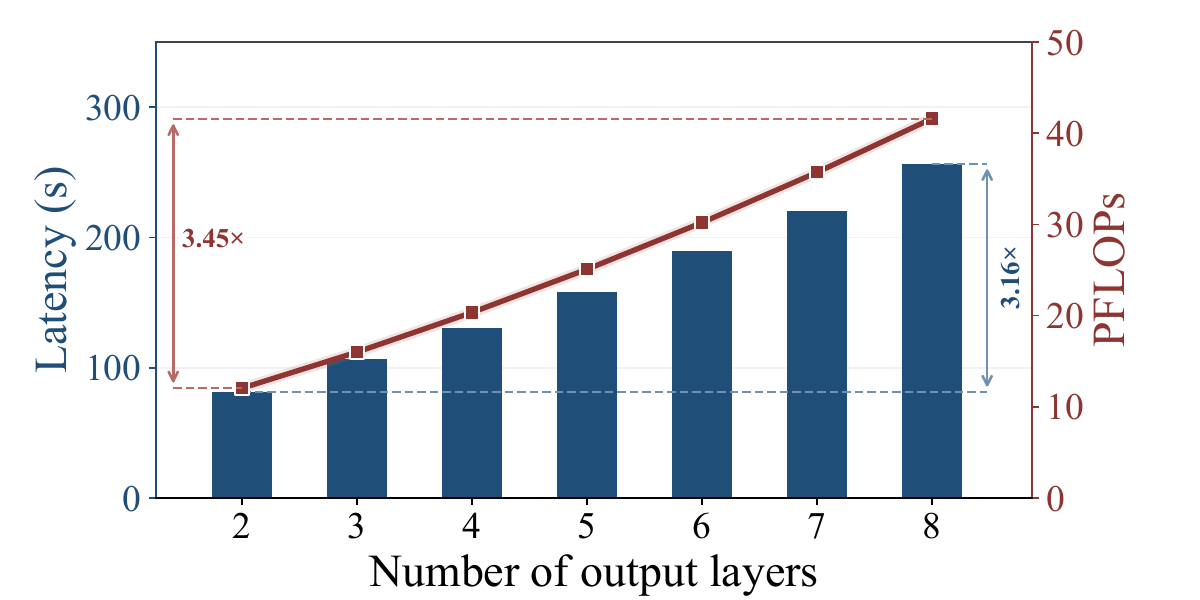}
            \caption{Inference Cost vs. Layer Count}
            \label{fig:motivation_a}
        \end{subfigure}

        \begin{subfigure}[b]{\linewidth}
            \centering
            \includegraphics[
                width=\linewidth,
                keepaspectratio
            ]{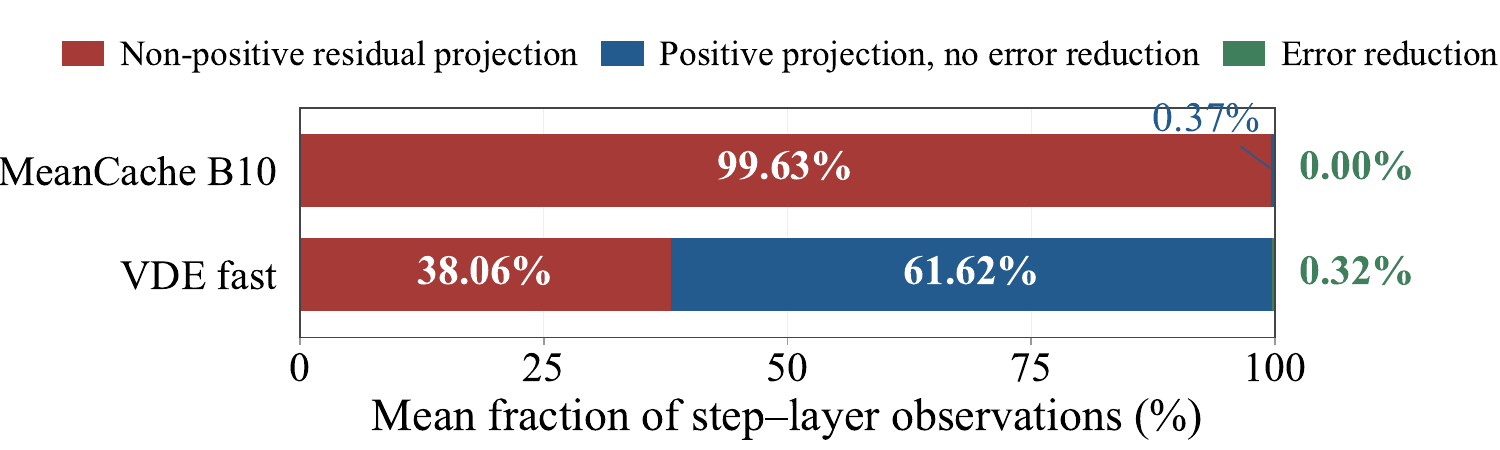}
            \caption{Local Correction Analysis}
            \label{fig:motivation_b}
        \end{subfigure}
    \end{minipage}
    \hfill
    \begin{minipage}[c]{0.50\textwidth}
        \centering

        \begin{subfigure}[b]{\linewidth}
            \centering
            \includegraphics[
                width=\linewidth,
                keepaspectratio
            ]{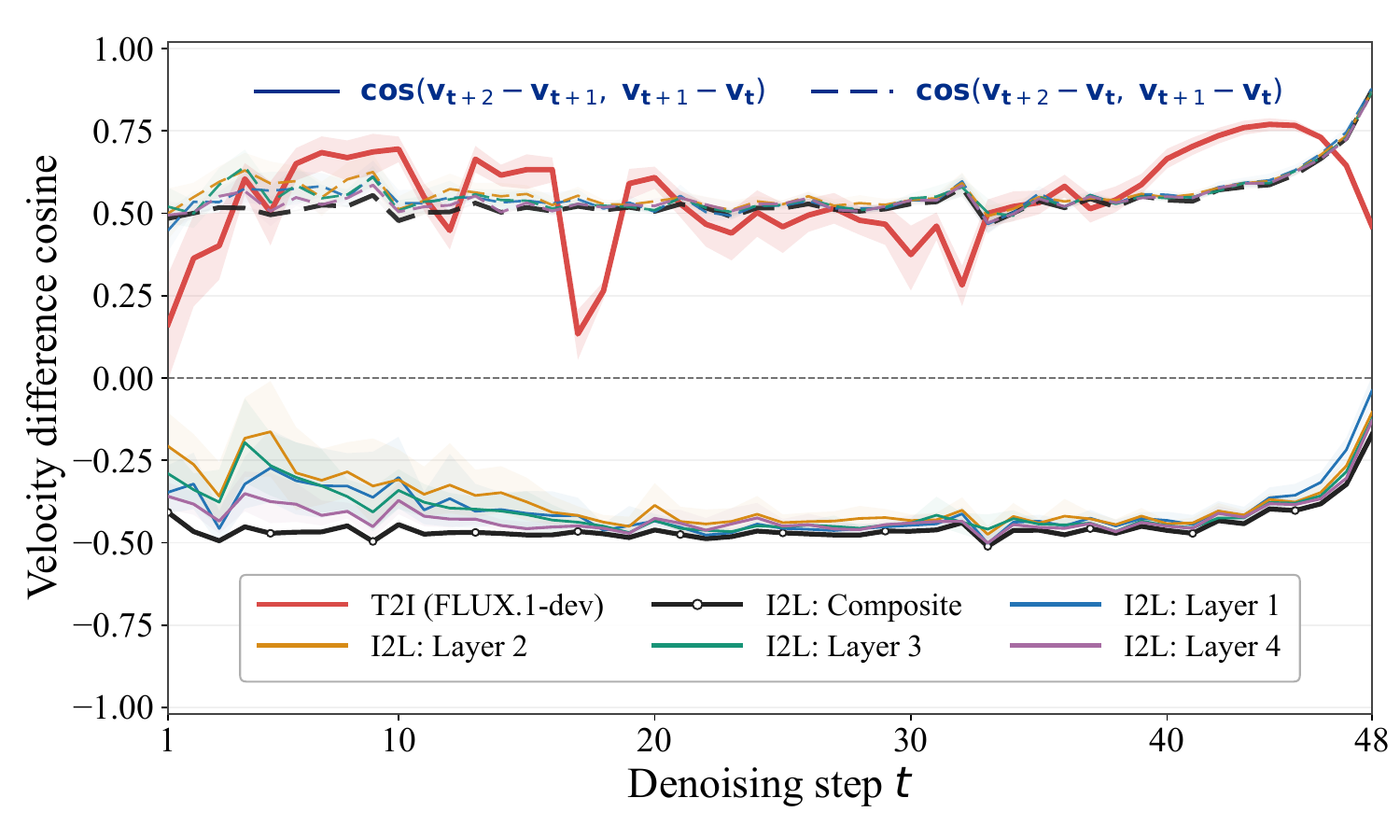}
            \caption{Velocity Increment Correlations}
            \label{fig:motivation_c}
        \end{subfigure}
    \end{minipage}

    \caption{\textbf{Inference cost and velocity prediction in multi-layer generation.} \textbf{(a)} The inference latency and computational cost of Qwen-Image-Layered increase with the number of output layers. The results are measured on a single NVIDIA A800 GPU. \textbf{(b)} Existing velocity forecasting methods often fail to improve the local displacement over direct velocity reuse. Most MeanCache corrections have no positive projection onto the direction needed to reduce the reuse error. Most VDE corrections have a positive projection but do not reduce the local displacement error. \textbf{(c)} Directional relationships between velocity increments in T2I and I2L. Solid lines show cosine similarity between consecutive increments. Dashed lines show cosine similarity between the second increment and their net change.}
    \label{fig:motivation}
\end{figure*}

\begin{abstract}

Recent advances in image-to-layer (I2L) generation enable editable image decomposition by jointly denoising multiple RGBA layers. However, the inference cost increases significantly with the number of output layers. Existing training-free velocity forecasting methods use historical velocities to accelerate Flow Matching inference. Surprisingly, when applied directly to I2L, they perform worse than simple velocity reuse. We further find that consecutive velocity increments in I2L are predominantly negatively correlated in direction. This makes velocity prediction difficult when simply extending historical trends. To address this challenge, we propose \textbf{FastLayered}, a training-free acceleration method based on Composite-Guided Velocity Calibration (CGVC). Specifically, we use the input-image latent as an approximate endpoint reference to predict the composite velocity. For each output layer, we combine velocities from the two most recent full model evaluations to predict its velocity direction. We then use the relative magnitude change of the composite velocity to predict its velocity magnitude. Experiments on Crello and Figma show that FastLayered achieves a \textbf{5.44$\times$} inference speedup on Qwen-Image-Layered while maintaining competitive generation quality. Under identical sampling schedules, FastLayered outperforms the compared velocity forecasting methods in both layer decomposition accuracy and global reconstruction fidelity. It reduces LPIPS by \textbf{up to 14.68\%} and improves PSNR by \textbf{up to 1.95 dB}.

\end{abstract}   



\section{Introduction}
\label{sec:intro}

Image-to-Layer (I2L) generation aims to decompose a single image into multiple RGBA layers \cite{liu2026ominpsd,nie2025decomposition,wang2026diffdecompose,jia2026pixelart,tang2026masked}. These layers allow users to independently edit, rearrange, and reuse image elements. Related work has also explored generating editable image layers directly~\cite{fontanella2024generating} and obtaining layers through generation followed by decomposition~\cite{kang2025layeringdiff}. Qwen-Image-Layered~\cite{yin2026qwen} enables editable image decomposition by jointly denoising multiple RGBA layers with Flow Matching. However, as the number of output layers increases, joint denoising incurs substantially higher inference latency and computational cost, as shown in Fig.~\ref{fig:motivation_a}. Achieving efficient multi-layer generation without sacrificing layer decomposition accuracy or reconstruction fidelity therefore remains challenging.

To accelerate Flow Matching inference, existing training-free velocity forecasting methods (e.g., MeanCache~\cite{gao2026meancache} and VDE~\cite{Tan_2026_CVPR}) use historical velocity information to predict updates at skipped steps~\cite{gao2026meancache,Tan_2026_CVPR}. \textbf{Surprisingly, when applied directly to I2L, these methods perform worse than simple velocity reuse in both layer decomposition accuracy and global reconstruction fidelity.} As shown in Fig.~\ref{fig:motivation_b}, we perform a local diagnosis along each method's own accelerated trajectory to analyze this performance gap. At each prediction step, we compute a reference displacement at the current state using a full model evaluation and a single Euler step. We then compare the displacements produced by velocity forecasting and direct velocity reuse with this reference. The vast majority of MeanCache~\cite{gao2026meancache} corrections have no positive component in the direction needed to reduce the error from velocity reuse. Most VDE~\cite{Tan_2026_CVPR} corrections have a positive component in that direction, but do not reduce the local displacement error. To further understand the difficulty of velocity prediction in I2L, we analyze how velocity increments evolve over time. We find predominantly positive directional correlations between consecutive velocity increments in text-to-image (T2I) generation. In I2L, however, these correlations are predominantly negative for the jointly generated composite-image latent and each output layer, as shown in Fig.~\ref{fig:motivation_c}. This observation suggests that velocity increments in I2L are \textbf{difficult to predict by simply extending historical trends}.

\begin{figure*}[t]
    \centering

    \begin{subfigure}[b]{0.48\textwidth}
        \centering
        \begin{minipage}[c][4.2cm][c]{\linewidth}
            \centering
            \includegraphics[
                width=\linewidth,
                height=4.2cm,
                keepaspectratio
            ]{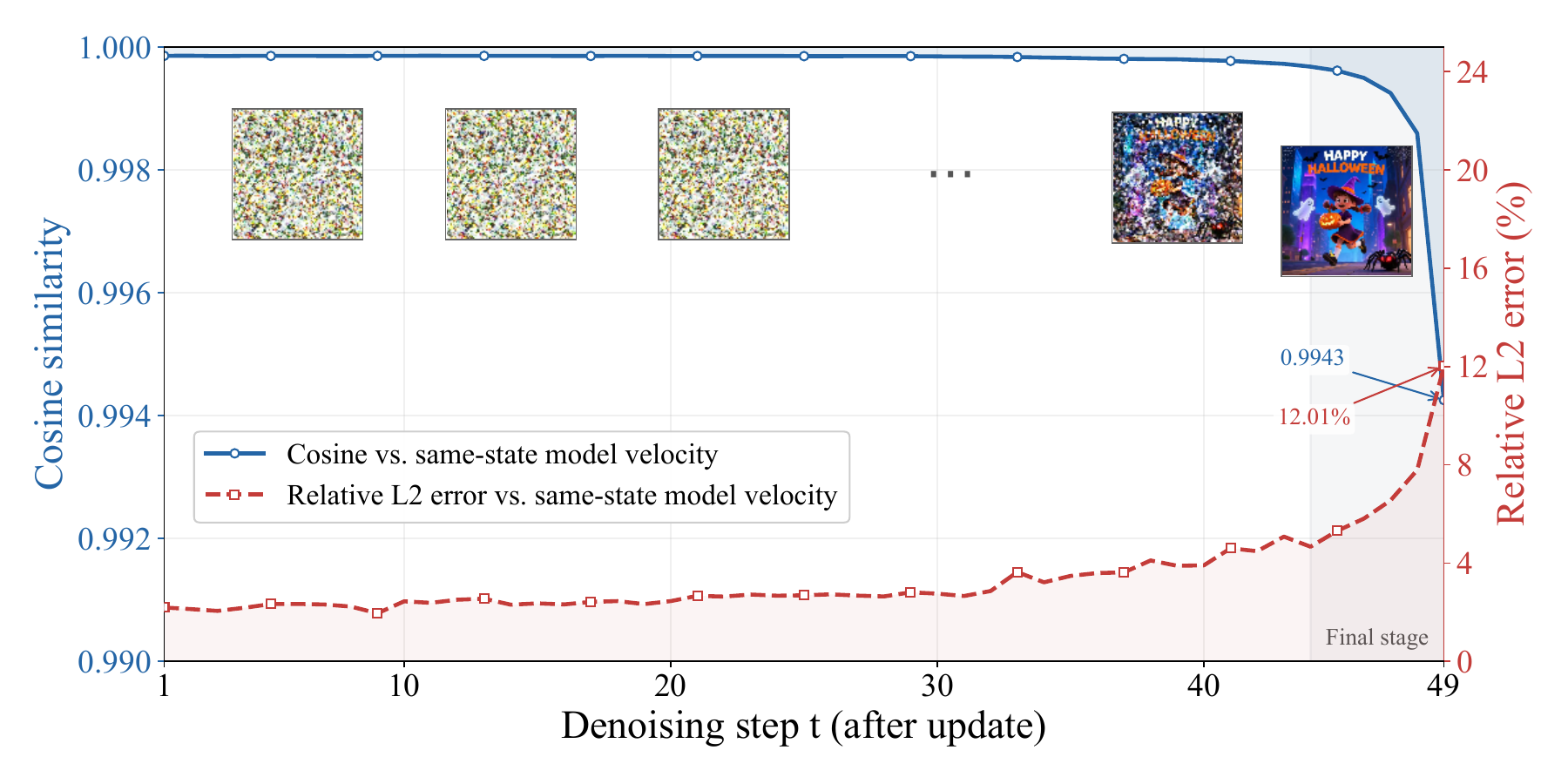}
        \end{minipage}
        \caption{Composite Velocity Consistency}
        \label{fig:cgvc_composite}
    \end{subfigure}
    \hfill
    \begin{subfigure}[b]{0.48\textwidth}
        \centering
        \begin{minipage}[c][4.2cm][c]{\linewidth}
            \centering
            \includegraphics[
                width=\linewidth,
                height=4.2cm,
                keepaspectratio
            ]{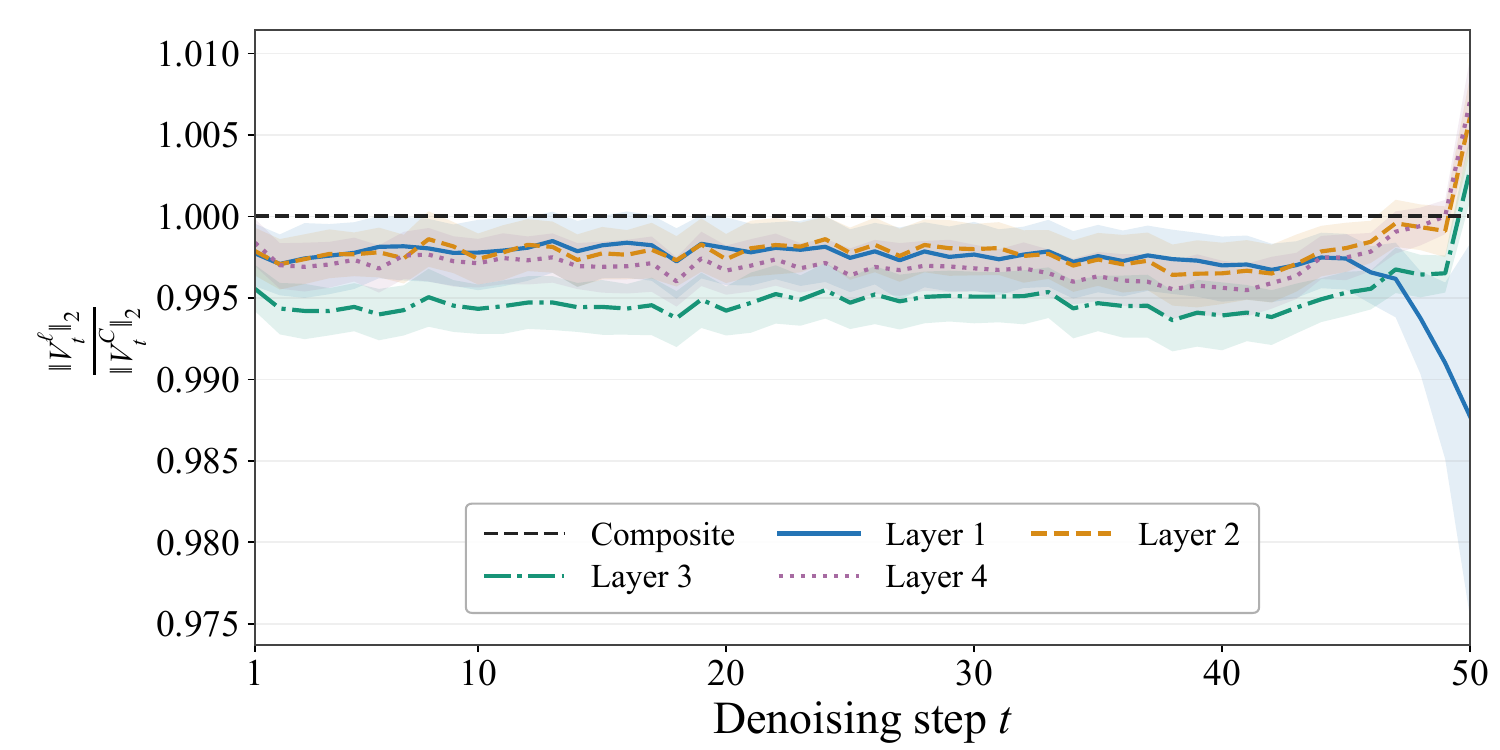}
        \end{minipage}
        \caption{Cross-Layer Velocity Magnitude Ratios}
        \label{fig:cgvc_layer}
    \end{subfigure}

    \caption{\textbf{Motivation for composite-guided velocity prediction.} \textbf{(a)} We construct a reference velocity from the updated composite latent using the input-image latent as an approximate endpoint. Its cosine similarity (blue) and relative L2 error (red) are measured against the composite velocity from a full model evaluation at the same state. The images show the decoded composite states after selected denoising updates. \textbf{(b)} The layer-to-composite velocity magnitude ratios remain relatively stable across most steps, suggesting that composite velocity changes can guide layer magnitude prediction. The black dashed line shows the composite-to-composite ratio.}
    \label{fig:cgvc_motivation}
\end{figure*}

To address these challenges, we propose \textbf{FastLayered}, a training-free acceleration method based on \textit{Composite-Guided Velocity Calibration (CGVC)}. CGVC predicts velocities for both the composite-image latent and the output layer latents jointly denoised by Qwen-Image-Layered. Specifically, CGVC uses the input-image latent as an approximate endpoint reference to predict the composite velocity. For each output layer latent, we combine its velocities from the two most recent full model evaluations to predict its velocity direction. We then calibrate its velocity magnitude using the relative magnitude change of the composite velocity.

\noindent\textbf{(1) Composite velocity prediction.} As shown in Fig.~\ref{fig:cgvc_composite}, we use the input-image latent as an approximate endpoint reference to construct a reference velocity from the updated composite-image latent. Compared with the composite velocity obtained from a full model evaluation at the same state, the reference velocity shows high directional consistency and a small relative error over most denoising steps. Motivated by this observation, we construct a target velocity for the composite using the input-image latent. We use this target velocity as the predicted composite velocity.

\noindent\textbf{(2) Output layer velocity prediction.} As shown in Fig.~\ref{fig:motivation_c}, consecutive velocity increments are predominantly negatively correlated in direction. However, the second of two consecutive increments is positively correlated in direction with the net velocity change over those two steps. Based on this observation, we sum the velocities from the two most recent full model evaluations for each output layer. We then normalize the resulting velocity and use it as the predicted velocity direction.



As shown in Fig.~\ref{fig:cgvc_layer}, the velocity magnitude ratios between each output layer and the composite remain relatively stable across most denoising steps. This observation suggests that the relative change in the composite velocity magnitude can guide velocity magnitude prediction for the output layers. Therefore, we compute a scaling factor as the ratio of the predicted composite velocity magnitude to the composite velocity magnitude from the most recent full model evaluation. We apply this factor to each output layer's velocity magnitude from the most recent full model evaluation to predict its velocity magnitude. Finally, we combine the predicted direction and magnitude to construct the predicted velocity for each output layer.

Extensive experiments on Crello~\cite{yamaguchi2021canvasvae} and Figma~\cite{yun2026redesign} show that FastLayered achieves a \textbf{5.44$\times$} inference speedup on Qwen-Image-Layered, while providing more accurate layer decomposition and higher-fidelity global reconstruction.

Our main contributions are summarized as follows:
\begin{itemize}
    \item We reveal the local correction difficulties faced by existing velocity forecasting methods in I2L inference. We further analyze the directional relationships between consecutive velocity increments and their net change. We also observe synchronized relative velocity magnitude changes between the composite and output layers. These observations support multi-layer velocity prediction.

    \item We propose \textbf{FastLayered}, a training-free acceleration method based on \textit{Composite-Guided Velocity Calibration (CGVC)}. CGVC uses the input-image latent to predict the composite velocity. For each output layer, it combines historical velocities to predict the direction and uses the relative magnitude change of the composite velocity to predict the magnitude.

    \item We conduct extensive experiments on Crello and Figma. FastLayered achieves a \textbf{5.44$\times$} inference speedup on Qwen-Image-Layered. Under identical sampling schedules, it outperforms the compared velocity forecasting methods in both layer decomposition accuracy and global reconstruction fidelity.
\end{itemize}
\section{Related Work}
\label{sec:related work}

\subsection{Image-to-Layer Generation}

Editable layer representations have become an active research area in transparent layer generation \cite{pu2025art,huang2025dreamlayer} and image decomposition \cite{yang2025generative,suzuki2025layerd,yin2026qwen}. They allow image elements to be modified independently. Early layer generation methods explored the use of generative models to synthesize transparent image content~\cite{zhang2023text2layer}. For example, LayerDiffuse~\cite{zhang2024transparent} introduces transparency into the latent space of diffusion models to generate transparent layers. LayerDiff~\cite{huang2024layerdiff} uses collaborative modeling across layers to generate coherent multi-layer images from text.

Another line of work decomposes an existing image into editable layers. Given an image and an object mask, LayerDecomp~\cite{yang2025generative} separates the image into a transparent foreground and a clean background while preserving visual effects such as shadows. LayerD~\cite{suzuki2025layerd} progressively recovers image elements by iteratively extracting foreground layers and completing the background. More recently, Qwen-Image-Layered~\cite{yin2026qwen} jointly denoises a variable number of RGBA layers to decompose an input image in an end-to-end framework. These semantically disentangled layers can be edited independently, providing a flexible representation for image editing. Furthermore, ReDesign~\cite{yun2026redesign} combines layer decomposition with other tools to recover editable design structures, including text, vector shapes, and hierarchical relationships. These works advance editable image representations. However, Qwen-Image-Layered repeatedly denoises multiple layers jointly during inference. Its computational cost increases with the number of output layers. Efficient inference is therefore important for the practical application of multi-layer image decomposition.

\subsection{Training-Free Acceleration Methods}

Existing training-free acceleration methods exploit computational redundancy between consecutive denoising steps to reduce repeated computation during inference \cite{ma2024deepcache,li2024faster,selvaraju2024fora,zhao2025real,kahatapitiya2025adaptive,Liu_2025_ICCV,liu2025timestep}. One line of work caches and reuses intermediate features or residuals, avoiding a full network computation at every step \cite{liu2024faster,chen2024delta,ICLR2025_bbe024e0,chu2025omnicache,zhou2025less,bu2026dicache}. TeaCache~\cite{liu2025timestep} estimates changes in model outputs using changes in inputs modulated by timestep embeddings. It then uses these estimates to adaptively decide when to reuse cached results. DiCache~\cite{bu2026dicache} uses information from shallow computations to detect changes during inference. It combines this information with historical caches to estimate the current result and adaptively select between caching and computation.

Another line of work uses historical information to estimate velocities and reduce the number of full model evaluations \cite{ICLR2026_d58e4a96,liu2026accelerating,gao2026meancache,Tan_2026_CVPR}. MeanCache~\cite{gao2026meancache} constructs velocity corrections from cached historical information to convert instantaneous velocities into interval-average velocities. VDE~\cite{Tan_2026_CVPR} decomposes the velocity into components relative to the current input. It then uses historical information to estimate these components and predict the current velocity. However, when these velocity estimation methods are applied to I2L, the directions and magnitudes of their corrections may not match the updates required at the current state. This mismatch can reduce velocity prediction accuracy.
\section{Method}

\subsection{Background}

Qwen-Image-Layered is an Image-to-Layer (I2L) generation model based on Flow Matching~\cite{jin2025pyramidal,albergo2022building}. It decomposes an input image into multiple editable RGBA layers through joint denoising. We denote the input-image latent by $z_I$. The joint latent state at sampling step $t$ is
\begin{equation}
    z_t = (z_t^C, z_t^1, \ldots, z_t^L),
    \label{eq:joint_latent_state}
\end{equation}
where $z_t^C$ is the composite-image latent, $z_t^\ell$ is the $\ell$-th output layer latent, and $L$ is the number of output layers. The input-image latent $z_I$ remains fixed throughout inference.

In Flow Matching, we consider a linear interpolation path between a clean state $x_0$ and a noise state $\epsilon$. The state at noise level $\sigma_t$ is
\begin{equation}
    x_t = (1-\sigma_t)x_0 + \sigma_t\epsilon,
    \quad \sigma_t \in [0,1].
    \label{eq:flow_interpolation}
\end{equation}
Here, $x_0$ is the zero-noise endpoint. The noise levels $\sigma_t=0$ and $\sigma_t=1$ correspond to the clean and noise endpoints, respectively. For $\sigma_t>0$, the reference velocity between the zero-noise endpoint and the current state on this path is
\begin{equation}
    v_{\mathrm{ref}}(x_t,\sigma_t)
    = \frac{x_t-x_0}{\sigma_t}.
    \label{eq:reference_velocity}
\end{equation}

During inference, the noise level decreases along the sampling schedule, and the state is updated using the Euler method. Given the current joint state $z_t$, the corresponding noise level $\sigma_t$, and the input condition, a full model evaluation jointly computes the velocities of the composite and all output layers
\begin{equation}
    v(z_t,t) = \bigl(v^C(z_t,t),v^1(z_t,t),\ldots,v^L(z_t,t)\bigr).
    \label{eq:joint_velocity}
\end{equation}
Here, $t$ denotes the current sampling step. The terms $v^C(z_t,t)$ and $v^\ell(z_t,t)$ denote the velocities of the composite latent and the $\ell$-th output layer latent at this step, respectively.

To describe our method, we denote three consecutive full-evaluation steps by $r$, $t$, and $s$, where $r<t<s$. The steps $r$ and $t$ are the two most recent full-evaluation steps, and $s$ is the next one. We only cache velocities from the two most recent full model evaluations. At each subsequent full-evaluation step, we replace the older cached velocities with the newly computed velocities.

\begin{figure*}[t]
    \centering
    \includegraphics[width=0.75\linewidth]{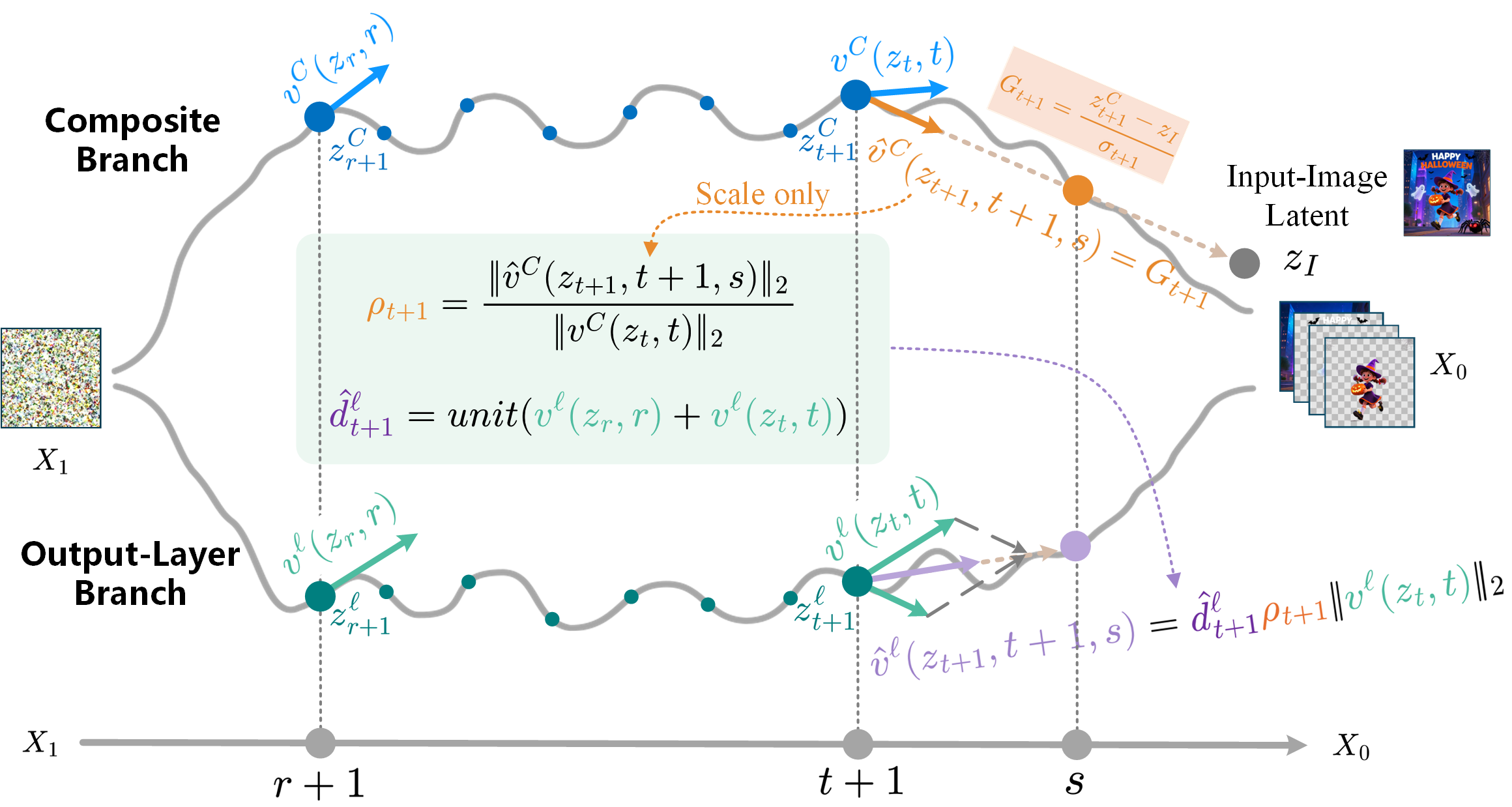}
    \caption{\textbf{Overview of the FastLayered framework.} $X_1$ and $X_0$ denote the initial noise state and the final generated result, respectively. The gray curves illustrate the denoising process. For the composite, we use the input-image latent $z_I$ as an approximate endpoint reference. We construct the predicted velocity $\widehat{v}^{C}(z_{t+1},t+1,s)$ from the updated composite latent. For each output layer, we combine the velocities from the two most recent full model evaluations to predict its velocity direction. We use the predicted composite velocity to construct a scaling factor $\rho_{t+1}$, which calibrates the velocity magnitude of each output layer. The historical velocities shown next to $z_{r+1}$ and $z_{t+1}$ are obtained from full model evaluations at $r$ and $t$, respectively, before the corresponding state updates.}
    \label{fig:overview}
\end{figure*}

\subsection{Composite Velocity Prediction}

The core of FastLayered is Composite-Guided Velocity Calibration (CGVC). CGVC predicts velocities for the composite and output layers separately. It also uses the predicted composite velocity to guide velocity magnitude calibration for the output layers. The overall framework is shown in Fig.~\ref{fig:overview}.

At full-evaluation step $t$, the model computes the velocities $v(z_t,t)$ for all branches from the current joint state $z_t$. We use these velocities to perform one Euler update and obtain the joint state $z_{t+1}$ at the next sampling step. Here, $z_{t+1}^C$ is the updated composite latent, and its corresponding noise level is $\sigma_{t+1}$.

The zero-noise endpoint of the composite is unknown during inference. We use the input-image latent as an approximate endpoint reference. As shown in Fig.~\ref{fig:cgvc_composite}, the resulting reference velocity is close to the composite velocity obtained from a full model evaluation at the same updated state. We construct the composite target velocity as
\begin{equation}
    G_{t+1} = \frac{z_{t+1}^C-z_I}{\sigma_{t+1}},
    \label{eq:composite_target_velocity}
\end{equation}
where $\sigma_{t+1}$ is the remaining noise span from the updated state to the zero-noise endpoint. This target velocity corresponds to a straight reference trajectory from the updated composite state toward the input-image latent. We use it as the predicted composite velocity, denoted by $\widehat{v}^C(z_{t+1},t+1,s)=G_{t+1}$.

Here, $s$ denotes the next full-evaluation step and identifies the subsequent update range for the predicted velocity. The magnitude of the predicted composite velocity also guides velocity magnitude prediction for the output layers. We describe this process in the next subsection.

\subsection{Output Layer Velocity Prediction}

For each output layer, we predict the velocity direction and magnitude separately. We then combine them to obtain the predicted velocity.

For velocity direction prediction, we sum the velocities of the $\ell$-th output layer from the two most recent full-evaluation steps $r$ and $t$. We then normalize the resulting velocity to obtain the predicted direction
\begin{equation}
    \widehat{d}_{t+1}^\ell
    = \frac{v^\ell(z_r,r)+v^\ell(z_t,t)}
    {\|v^\ell(z_r,r)+v^\ell(z_t,t)\|_2},
    \label{eq:layer_velocity_direction}
\end{equation}
where $\widehat{d}_{t+1}^\ell$ is the predicted unit velocity direction for the $\ell$-th output layer.

For velocity magnitude prediction, we use the predicted composite velocity to construct a scaling factor. Specifically, we compute the ratio of the magnitude of the predicted composite velocity $\widehat{v}^C(z_{t+1},t+1,s)$ to that of the composite velocity $v^C(z_t,t)$ from the most recent full model evaluation
\begin{equation}
    \rho_{t+1} = \frac{\|\widehat{v}^C(z_{t+1},t+1,s)\|_2}
    {\|v^C(z_t,t)\|_2},
    \label{eq:shared_scaling_factor}
\end{equation}

\begin{table*}[t]
\caption{Comparison of acceleration methods for Qwen-Image-Layered on the Crello and Figma datasets. We report inference efficiency, layer decomposition accuracy, and global reconstruction fidelity. `Speed.'\ and `Lat.'\ denote the speedup relative to vanilla inference with $T=50$ and the per-image inference latency in seconds, respectively. The values 0--5 under RGB L1 and Alpha soft IoU indicate the maximum allowed number of layer merges. $T$ denotes the number of sampling steps for vanilla inference, and $\mathcal{B}$ denotes the number of full model evaluations for MeanCache. FastLayered performs \textbf{\textit{9 full model evaluations}}. These include evaluations at the first two steps and the final step. Between them, we use a fixed-interval schedule. Among the acceleration methods, the best results are shown in \textbf{bold}, and the second-best results are \underline{underlined}.}
\label{tab:quantitative_comparison}
\centering
\resizebox{\textwidth}{!}{%
\begin{tabular}{ll*{16}{c}}
\toprule


\multicolumn{2}{c}{\multirow{3}{*}{\textbf{Methods}}}
& \multicolumn{2}{c}{\textbf{Efficiency}}
& \multicolumn{12}{c}{\textbf{Decomposition-level}}
& \multicolumn{2}{c}{\textbf{Global-level}} \\

\cmidrule(lr){3-4}
\cmidrule(lr){5-16}
\cmidrule(l){17-18}

\multicolumn{2}{c}{}
& \multirow{2}{*}{\textbf{Speed.} $\uparrow$}
& \multirow{2}{*}{\textbf{Lat. (s)} $\downarrow$}
& \multicolumn{6}{c}{\textbf{RGB L1} $\downarrow$}
& \multicolumn{6}{c}{\textbf{Alpha soft IoU} $\uparrow$}
& \multirow{2}{*}{\textbf{PSNR} $\uparrow$}
& \multirow{2}{*}{\textbf{LPIPS} $\downarrow$} \\

\cmidrule(lr){5-10}
\cmidrule(lr){11-16}

\multicolumn{2}{c}{}
& & &
\textbf{0} & \textbf{1} & \textbf{2}
& \textbf{3} & \textbf{4} & \textbf{5}
& \textbf{0} & \textbf{1} & \textbf{2}
& \textbf{3} & \textbf{4} & \textbf{5}
& & \\

\midrule


\multicolumn{18}{c}{\textit{Crello Dataset}} \\


\multirow{5}{*}{Vanilla~\cite{yin2026qwen}}
& $T=50$
& 1.00$\times$ & 128.59
& 0.1973 & 0.1570 & 0.1192 & 0.1058 & 0.1005 & 0.0989
& 0.5957 & 0.7078 & 0.8000 & 0.8443 & 0.8606 & 0.8649
& 27.57 & 0.0712 \\

& $T=25$
& 1.99$\times$ & 64.48
& 0.2028 & 0.1606 & 0.1214 & 0.1081 & 0.1027 & 0.1011
& 0.5928 & 0.7036 & 0.7953 & 0.8400 & 0.8554 & 0.8594
& 27.25 & 0.0776 \\

& $T=20$
& 2.49$\times$ & 51.63
& 0.2044 & 0.1621 & 0.1218 & 0.1084 & 0.1032 & 0.1016
& 0.5922 & 0.7021 & 0.7934 & 0.8375 & 0.8530 & 0.8571
& 27.18 & 0.0799 \\

& $T=15$
& 3.31$\times$ & 38.81
& 0.2066 & 0.1637 & 0.1232 & 0.1098 & 0.1045 & 0.1027
& 0.5883 & 0.6970 & 0.7876 & 0.8312 & 0.8467 & 0.8511
& 27.00 & 0.0845 \\

& $T=10$
& \underline{4.95}$\times$ & \underline{25.98}
& 0.2094 & 0.1662 & 0.1262 & 0.1119 & 0.1066 & 0.1047
& 0.5809 & 0.6863 & 0.7742 & 0.8167 & 0.8323 & 0.8369
& 26.50 & 0.0984 \\

\hdashline[0.8pt/1.5pt]


\multicolumn{2}{l}{TeaCache ($\theta=0.2$)~\cite{liu2025timestep}}
& 2.87$\times$ & 44.87
& \underline{0.1972} & \underline{0.1571}
& \underline{0.1196} & \underline{0.1057}
& \underline{0.1006} & \underline{0.0990}
& 0.5960 & \underline{0.7080}
& \underline{0.8003} & \underline{0.8449}
& \underline{0.8604} & 0.8644
& \underline{27.53} & \underline{0.0715} \\

\multicolumn{2}{l}{TeaCache ($\theta=0.3$)~\cite{liu2025timestep}}
& 3.52$\times$ & 36.48
& \underline{0.1972} & 0.1574
& \textbf{0.1194} & 0.1058 & 0.1009 & 0.0991
& 0.5971 & 0.7075 & 0.7999 & 0.8446 & 0.8600
& \underline{0.8645}
& 27.49 & 0.0722 \\


\multicolumn{2}{l}{DiCache ($\delta=0.4$)~\cite{bu2026dicache}}
& 3.17$\times$ & 40.60
& 0.2019 & 0.1613 & 0.1251 & 0.1119 & 0.1066 & 0.1049
& \underline{0.5998} & 0.7060 & 0.7948 & 0.8390 & 0.8543 & 0.8584
& 26.43 & 0.0825 \\


\multicolumn{2}{l}{VDE-slow~\cite{Tan_2026_CVPR}}
& 2.08$\times$ & 61.92
& 0.2035 & 0.1634 & 0.1253 & 0.1131 & 0.1076 & 0.1060
& 0.5884 & 0.6966 & 0.7864 & 0.8304 & 0.8461 & 0.8501
& 25.81 & 0.0775 \\

\multicolumn{2}{l}{VDE-fast~\cite{Tan_2026_CVPR}}
& 2.62$\times$ & 49.11
& 0.2057 & 0.1647 & 0.1258 & 0.1135 & 0.1081 & 0.1064
& 0.5895 & 0.6988 & 0.7893 & 0.8333 & 0.8491 & 0.8530
& 25.79 & 0.0790 \\


\multicolumn{2}{l}{MeanCache ($\mathcal{B}=25$)~\cite{gao2026meancache}}
& 1.99$\times$ & 64.50
& 0.2054 & 0.1624 & 0.1223 & 0.1088 & 0.1036 & 0.1019
& 0.5910 & 0.7017 & 0.7930 & 0.8381 & 0.8537 & 0.8577
& 27.28 & 0.0807 \\

\multicolumn{2}{l}{MeanCache ($\mathcal{B}=17$)~\cite{gao2026meancache}}
& 2.93$\times$ & 43.94
& 0.2039 & 0.1612 & 0.1210 & 0.1076 & 0.1025 & 0.1008
& 0.5917 & 0.7029 & 0.7946 & 0.8398 & 0.8554 & 0.8592
& 27.30 & 0.0798 \\

\multicolumn{2}{l}{MeanCache ($\mathcal{B}=10$)~\cite{gao2026meancache}}
& \underline{4.95}$\times$ & \underline{25.98}
& 0.2084 & 0.1648 & 0.1257 & 0.1128 & 0.1075 & 0.1057
& 0.5911 & 0.6991 & 0.7889 & 0.8332 & 0.8482 & 0.8524
& 27.07 & 0.0824 \\


\rowcolor{gray!15}
\multicolumn{2}{l}{\textbf{FastLayered}}
& \textbf{5.44}$\times$ & \textbf{23.66}
& \textbf{0.1960} & \textbf{0.1569} & \textbf{0.1194}
& \textbf{0.1049} & \textbf{0.1000} & \textbf{0.0983}
& \textbf{0.6010} & \textbf{0.7150} & \textbf{0.8100}
& \textbf{0.8531} & \textbf{0.8684} & \textbf{0.8733}
& \textbf{27.70} & \textbf{0.0707} \\

\midrule


\multicolumn{18}{c}{\textit{Figma Dataset}} \\


\multirow{5}{*}{Vanilla~\cite{yin2026qwen}}
& $T=50$
& $1.00\times$ & 128.66
& 0.2205 & 0.1900 & 0.1655 & 0.1463 & 0.1305 & 0.1225
& 0.4172 & 0.4929 & 0.5650 & 0.6230 & 0.6684 & 0.7038
& 25.60 & 0.1135 \\

& $T=25$
& $1.99\times$ & 64.68
& 0.2237 & 0.1906 & 0.1655 & 0.1471 & 0.1323 & 0.1247
& 0.4159 & 0.4894 & 0.5619 & 0.6192 & 0.6658 & 0.6984
& 25.47 & 0.1181 \\

& $T=20$
& $2.48\times$ & 51.82
& 0.2243 & 0.1913 & \underline{0.1648}
& 0.1468 & 0.1330 & 0.1247
& 0.4149 & 0.4886 & 0.5603 & 0.6176 & 0.6634 & 0.6962
& 25.42 & 0.1203 \\

& $T=15$
& $3.30\times$ & 38.94
& 0.2255 & 0.1918 & 0.1652 & 0.1475 & 0.1335 & 0.1259
& 0.4128 & 0.4854 & 0.5567 & 0.6151 & 0.6598 & 0.6924
& 25.29 & 0.1240 \\

& $T=10$
& $4.95\times$ & 25.99
& 0.2263 & 0.1934 & 0.1664 & 0.1483 & 0.1350 & 0.1269
& 0.4102 & 0.4818 & 0.5505 & 0.6087 & 0.6537 & 0.6864
& 24.98 & 0.1349 \\

\hdashline[0.8pt/1.5pt]


\multicolumn{2}{l}{TeaCache ($\theta=0.2$)~\cite{liu2025timestep}}
& $2.87\times$ & 44.87
& 0.2202 & 0.1904 & 0.1656
& \underline{0.1462} & \textbf{0.1319} & \textbf{0.1238}
& 0.4172 & \underline{0.4931}
& \underline{0.5652} & \underline{0.6229}
& \underline{0.6698} & \underline{0.7020}
& \textbf{25.59} & \textbf{0.1145} \\

\multicolumn{2}{l}{TeaCache ($\theta=0.3$)~\cite{liu2025timestep}}
& $3.52\times$ & 36.51
& \underline{0.2200} & \underline{0.1900}
& 0.1650 & 0.1468
& \underline{0.1321} & 0.1247
& 0.4187 & 0.4921 & 0.5640 & 0.6218 & 0.6686 & 0.7014
& 25.45 & \underline{0.1151} \\


\multicolumn{2}{l}{DiCache ($\delta=0.4$)~\cite{bu2026dicache}}
& $3.16\times$ & 40.69
& 0.2252 & 0.1953 & 0.1699 & 0.1524 & 0.1387 & 0.1281
& \underline{0.4199} & 0.4926 & 0.5620 & 0.6212 & 0.6680 & 0.7011
& 24.53 & 0.1269 \\


\multicolumn{2}{l}{VDE-slow~\cite{Tan_2026_CVPR}}
& $2.08\times$ & 61.97
& 0.2245 & 0.1937 & 0.1688 & 0.1510 & 0.1372 & 0.1287
& 0.4141 & 0.4873 & 0.5580 & 0.6159 & 0.6617 & 0.6957
& 24.83 & 0.1164 \\

\multicolumn{2}{l}{VDE-fast~\cite{Tan_2026_CVPR}}
& $2.62\times$ & 49.15
& 0.2259 & 0.1950 & 0.1696 & 0.1515 & 0.1377 & 0.1293
& 0.4137 & 0.4878 & 0.5579 & 0.6161 & 0.6622 & 0.6951
& 24.73 & 0.1192 \\


\multicolumn{2}{l}{MeanCache ($\mathcal{B}=25$)~\cite{gao2026meancache}}
& $1.99\times$ & 64.51
& 0.2256 & 0.1927 & 0.1677 & 0.1490 & 0.1350 & 0.1263
& 0.4145 & 0.4894 & 0.5600 & 0.6174 & 0.6634 & 0.6981
& 25.28 & 0.1241 \\

\multicolumn{2}{l}{MeanCache ($\mathcal{B}=17$)~\cite{gao2026meancache}}
& $2.93\times$ & 43.98
& 0.2246 & 0.1920 & 0.1665 & 0.1481 & 0.1345 & 0.1257
& 0.4157 & 0.4897 & 0.5608 & 0.6184 & 0.6657 & 0.7003
& 25.32 & 0.1230 \\

\multicolumn{2}{l}{MeanCache ($\mathcal{B}=10$)~\cite{gao2026meancache}}
& \underline{4.95}$\times$ & \underline{26.00}
& 0.2274 & 0.1942 & 0.1681 & 0.1499 & 0.1363 & 0.1290
& 0.4151 & 0.4880 & 0.5592 & 0.6160 & 0.6614 & 0.6939
& 25.03 & 0.1277 \\


\rowcolor{gray!15}
\multicolumn{2}{l}{\textbf{FastLayered}}
& \textbf{5.44}$\times$ & \textbf{23.62}
& \textbf{0.2184} & \textbf{0.1885} & \textbf{0.1641}
& \textbf{0.1456} & \underline{0.1321} & \underline{0.1245}
& \textbf{0.4214} & \textbf{0.4940} & \textbf{0.5686}
& \textbf{0.6256} & \textbf{0.6732} & \textbf{0.7050}
& \underline{25.48} & \textbf{0.1145} \\

\bottomrule
\end{tabular}%
}
\end{table*}

where $\rho_{t+1}$ is the shared scaling factor for calibrating the velocity magnitudes of the output layers.

We then apply $\rho_{t+1}$ to the velocity magnitude of each output layer from the most recent full model evaluation. The predicted velocity magnitude for the $\ell$-th output layer is
\begin{equation}
    m_{t+1}^\ell = \rho_{t+1}\|v^\ell(z_t,t)\|_2.
    \label{eq:layer_velocity_magnitude}
\end{equation}

Finally, we combine the predicted velocity direction and magnitude to obtain the predicted velocity for the $\ell$-th output layer
\begin{equation}
    \widehat{v}^\ell(z_{t+1},t+1,s) = m_{t+1}^\ell\widehat{d}_{t+1}^\ell.
    \label{eq:predicted_layer_velocity}
\end{equation}

The pseudocode for the complete FastLayered inference process is provided in the supplementary material.

\section{Experiments}

\subsection{Experimental Settings}

\noindent\textbf{Base Model and Datasets.}
We use Qwen-Image-Layered as the base image-to-layer generation model. We evaluate FastLayered on the Crello dataset~\cite{yamaguchi2021canvasvae} and the Figma dataset~\cite{yun2026redesign} introduced by ReDesign. The Crello and Figma test sets contain 1,954 and 909 images, respectively. In the main experiments, the model generates four RGBA layers for each input image.

\noindent\textbf{Comparison Methods.}
We compare FastLayered with vanilla inference using different numbers of sampling steps and existing training-free acceleration methods. The feature caching methods include TeaCache and DiCache. The velocity forecasting methods include VDE and MeanCache. We adapt the official implementations of the existing training-free acceleration methods and use their default configurations.

\noindent\textbf{Evaluation Metrics.}
We evaluate inference efficiency, layer decomposition, and global reconstruction. For efficiency, we report latency and speedup relative to $T=50$. For layer decomposition, we follow LayerD to align predicted and ground-truth layers in order, allowing adjacent layer merges. We report RGB L1, weighted by ground-truth alpha, and Alpha soft IoU for maximum merge counts from 0 to 5. For global reconstruction, we follow ReDesign and use PSNR~\cite{wang2004image} and LPIPS~\cite{zhang2018unreasonable} to compare the input image with the reconstruction rendered from the output layers.

\begin{table*}[t]
\caption{Comparison of FastLayered with existing velocity forecasting methods under the same sampling schedules on the Crello and Figma datasets.}
\label{tab:quantitative_comparison_CGVC}
\centering
\resizebox{\textwidth}{!}{%
\begin{tabular}{ll*{16}{c}}
\toprule

\multirow{3}{*}{\textbf{Sampling Schedule}}
& \multirow{3}{*}{\textbf{Methods}}
& \multicolumn{2}{c}{\textbf{Efficiency}}
& \multicolumn{12}{c}{\textbf{Decomposition-level}}
& \multicolumn{2}{c}{\textbf{Global-level}} \\

\cmidrule(lr){3-4}
\cmidrule(lr){5-16}
\cmidrule(l){17-18}

& &
\multirow{2}{*}{\textbf{Speed.} $\uparrow$}
& \multirow{2}{*}{\textbf{Lat. (s)} $\downarrow$}
& \multicolumn{6}{c}{\textbf{RGB L1} $\downarrow$}
& \multicolumn{6}{c}{\textbf{Alpha soft IoU} $\uparrow$}
& \multirow{2}{*}{\textbf{PSNR} $\uparrow$}
& \multirow{2}{*}{\textbf{LPIPS} $\downarrow$} \\

\cmidrule(lr){5-10}
\cmidrule(lr){11-16}

& & & &
0 & 1 & 2 & 3 & 4 & 5 &
0 & 1 & 2 & 3 & 4 & 5 &
& \\

\midrule

\multicolumn{18}{c}{\textit{Crello Dataset}} \\


& VDE~\cite{Tan_2026_CVPR}
& \textbf{2.08}$\times$ & \textbf{61.92}
& 0.2035 & 0.1634 & 0.1253 & 0.1131 & 0.1076 & 0.1060
& 0.5884 & 0.6966 & 0.7864 & 0.8304 & 0.8461 & 0.8501
& 25.81 & 0.0775 \\

\rowcolor{gray!15}
\cellcolor{white}\multirow{-2}{*}{VDE-slow}
& \textbf{FastLayered}
& 2.06$\times$ & 62.46
& \textbf{0.1958} & \textbf{0.1562} & \textbf{0.1183}
& \textbf{0.1048} & \textbf{0.0995} & \textbf{0.0978}
& \textbf{0.6000} & \textbf{0.7147} & \textbf{0.8090}
& \textbf{0.8537} & \textbf{0.8687} & \textbf{0.8732}
& \textbf{27.74} & \textbf{0.0703} \\

\hdashline[0.8pt/1.5pt]


& VDE~\cite{Tan_2026_CVPR}
& \textbf{2.62}$\times$ & \textbf{49.11}
& 0.2057 & 0.1647 & 0.1258 & 0.1135 & 0.1081 & 0.1064
& 0.5895 & 0.6988 & 0.7893 & 0.8333 & 0.8491 & 0.8530
& 25.79 & 0.0790 \\

\rowcolor{gray!15}
\cellcolor{white}\multirow{-2}{*}{VDE-fast}
& \textbf{FastLayered}
& 2.60$\times$ & 49.48
& \textbf{0.1957} & \textbf{0.1563} & \textbf{0.1187}
& \textbf{0.1054} & \textbf{0.1001} & \textbf{0.0984}
& \textbf{0.6002} & \textbf{0.7149} & \textbf{0.8097}
& \textbf{0.8545} & \textbf{0.8697} & \textbf{0.8741}
& \textbf{27.74} & \textbf{0.0703} \\

\hdashline[0.8pt/1.5pt]


& MeanCache~\cite{gao2026meancache}
& \textbf{1.99}$\times$ & \textbf{64.50}
& 0.2054 & 0.1624 & 0.1223 & 0.1088 & 0.1036 & 0.1019
& 0.5910 & 0.7017 & 0.7930 & 0.8381 & 0.8537 & 0.8577
& 27.28 & 0.0807 \\

\rowcolor{gray!15}
\cellcolor{white}\multirow{-2}{*}{MeanCache ($\mathcal{B}=25$)}
& \textbf{FastLayered}
& 1.98$\times$ & 64.95
& \textbf{0.1958} & \textbf{0.1563} & \textbf{0.1188}
& \textbf{0.1055} & \textbf{0.1003} & \textbf{0.0986}
& \textbf{0.5995} & \textbf{0.7137} & \textbf{0.8081}
& \textbf{0.8529} & \textbf{0.8687} & \textbf{0.8731}
& \textbf{27.64} & \textbf{0.0706} \\

\hdashline[0.8pt/1.5pt]


& MeanCache~\cite{gao2026meancache}
& \textbf{2.93}$\times$ & \textbf{43.94}
& 0.2039 & 0.1612 & 0.1210 & 0.1076 & 0.1025 & 0.1008
& 0.5917 & 0.7029 & 0.7946 & 0.8398 & 0.8554 & 0.8592
& 27.30 & 0.0798 \\

\rowcolor{gray!15}
\cellcolor{white}\multirow{-2}{*}{MeanCache ($\mathcal{B}=17$)}
& \textbf{FastLayered}
& 2.90$\times$ & 44.30
& \textbf{0.1953} & \textbf{0.1559} & \textbf{0.1185}
& \textbf{0.1049} & \textbf{0.0998} & \textbf{0.0980}
& \textbf{0.6007} & \textbf{0.7147} & \textbf{0.8101}
& \textbf{0.8547} & \textbf{0.8704} & \textbf{0.8749}
& \textbf{27.71} & \textbf{0.0700} \\

\hdashline[0.8pt/1.5pt]


& MeanCache~\cite{gao2026meancache}
& \textbf{4.95}$\times$ & \textbf{25.98}
& 0.2084 & 0.1648 & 0.1257 & 0.1128 & 0.1075 & 0.1057
& 0.5911 & 0.6991 & 0.7889 & 0.8332 & 0.8482 & 0.8524
& 27.07 & 0.0824 \\

\rowcolor{gray!15}
\cellcolor{white}\multirow{-2}{*}{MeanCache ($\mathcal{B}=10$)}
& \textbf{FastLayered}
& 4.90$\times$ & 26.25
& \textbf{0.1956} & \textbf{0.1569} & \textbf{0.1192}
& \textbf{0.1054} & \textbf{0.1004} & \textbf{0.0986}
& \textbf{0.6018} & \textbf{0.7157} & \textbf{0.8100}
& \textbf{0.8534} & \textbf{0.8690} & \textbf{0.8738}
& \textbf{27.73} & \textbf{0.0703} \\

\midrule
\multicolumn{18}{c}{\textit{Figma Dataset}} \\


& VDE~\cite{Tan_2026_CVPR}
& \textbf{2.08}$\times$ & \textbf{61.97}
& 0.2245 & 0.1937 & 0.1688 & 0.1510 & 0.1372 & 0.1287
& 0.4141 & 0.4873 & 0.5580 & 0.6159 & 0.6617 & 0.6957
& 24.83 & 0.1164 \\

\rowcolor{gray!15}
\cellcolor{white}\multirow{-2}{*}{VDE-slow}
& \textbf{FastLayered}
& 2.06$\times$ & 62.43
& \textbf{0.2206} & \textbf{0.1897} & \textbf{0.1662}
& \textbf{0.1464} & \textbf{0.1315} & \textbf{0.1230}
& \textbf{0.4194} & \textbf{0.4945} & \textbf{0.5681}
& \textbf{0.6269} & \textbf{0.6728} & \textbf{0.7071}
& \textbf{25.62} & \textbf{0.1130} \\

\hdashline[0.8pt/1.5pt]


& VDE~\cite{Tan_2026_CVPR}
& \textbf{2.62}$\times$ & \textbf{49.15}
& 0.2259 & 0.1950 & 0.1696 & 0.1515 & 0.1377 & 0.1293
& 0.4137 & 0.4878 & 0.5579 & 0.6161 & 0.6622 & 0.6951
& 24.73 & 0.1192 \\

\rowcolor{gray!15}
\cellcolor{white}\multirow{-2}{*}{VDE-fast}
& \textbf{FastLayered}
& 2.60$\times$ & 49.42
& \textbf{0.2205} & \textbf{0.1907} & \textbf{0.1654}
& \textbf{0.1450} & \textbf{0.1311} & \textbf{0.1228}
& \textbf{0.4190} & \textbf{0.4950} & \textbf{0.5681}
& \textbf{0.6269} & \textbf{0.6731} & \textbf{0.7071}
& \textbf{25.61} & \textbf{0.1129} \\

\hdashline[0.8pt/1.5pt]


& MeanCache~\cite{gao2026meancache}
& \textbf{1.99}$\times$ & \textbf{64.51}
& 0.2256 & 0.1927 & 0.1677 & 0.1490 & 0.1350 & 0.1263
& 0.4145 & 0.4894 & 0.5600 & 0.6174 & 0.6634 & 0.6981
& 25.28 & 0.1241 \\

\rowcolor{gray!15}
\cellcolor{white}\multirow{-2}{*}{MeanCache ($\mathcal{B}=25$)}
& \textbf{FastLayered}
& 1.98$\times$ & 64.82
& \textbf{0.2199} & \textbf{0.1895} & \textbf{0.1661}
& \textbf{0.1466} & \textbf{0.1310} & \textbf{0.1232}
& \textbf{0.4180} & \textbf{0.4944} & \textbf{0.5676}
& \textbf{0.6272} & \textbf{0.6739} & \textbf{0.7069}
& \textbf{25.57} & \textbf{0.1130} \\

\hdashline[0.8pt/1.5pt]


& MeanCache~\cite{gao2026meancache}
& \textbf{2.93}$\times$ & \textbf{43.98}
& 0.2246 & 0.1920 & 0.1665 & 0.1481 & 0.1345 & 0.1257
& 0.4157 & 0.4897 & 0.5608 & 0.6184 & 0.6657 & 0.7003
& 25.32 & 0.1230 \\

\rowcolor{gray!15}
\cellcolor{white}\multirow{-2}{*}{MeanCache ($\mathcal{B}=17$)}
& \textbf{FastLayered}
& 2.91$\times$ & 44.24
& \textbf{0.2203} & \textbf{0.1898} & \textbf{0.1650}
& \textbf{0.1460} & \textbf{0.1318} & \textbf{0.1231}
& \textbf{0.4182} & \textbf{0.4949} & \textbf{0.5691}
& \textbf{0.6270} & \textbf{0.6743} & \textbf{0.7076}
& \textbf{25.56} & \textbf{0.1134} \\

\hdashline[0.8pt/1.5pt]


& MeanCache~\cite{gao2026meancache}
& \textbf{4.95}$\times$ & \textbf{26.00}
& 0.2274 & 0.1942 & 0.1681 & 0.1499 & 0.1363 & 0.1290
& 0.4151 & 0.4880 & 0.5592 & 0.6160 & 0.6614 & 0.6939
& 25.03 & 0.1277 \\

\rowcolor{gray!15}
\cellcolor{white}\multirow{-2}{*}{MeanCache ($\mathcal{B}=10$)}
& \textbf{FastLayered}
& 4.91$\times$ & 26.22
& \textbf{0.2199} & \textbf{0.1890} & \textbf{0.1652}
& \textbf{0.1462} & \textbf{0.1328} & \textbf{0.1245}
& \textbf{0.4194} & \textbf{0.4945} & \textbf{0.5680}
& \textbf{0.6272} & \textbf{0.6725} & \textbf{0.7036}
& \textbf{25.45} & \textbf{0.1150} \\

\bottomrule
\end{tabular}%
}
\end{table*}

\noindent\textbf{Implementation Details.}
For each dataset, all methods are evaluated on the same test images using the same random seed and hardware. We set the resolution parameter to 640 for all experiments. In the main comparison in Table~\ref{tab:quantitative_comparison}, FastLayered performs full model evaluations at the first two steps and the final step. It uses a fixed-interval schedule for the intermediate steps. All experiments are conducted on a single NVIDIA A800 GPU. For more implementation details, please refer to the supplementary material.

\begin{table}[t]
\caption{Comparison of FastLayered with existing velocity forecasting methods under the same sampling schedules for different numbers of output layers on Figma. \# Layers denotes the number of output layers. The superscript avg indicates that the metric is averaged over maximum allowed layer merge counts from 0 to 5. Detailed results are provided in the supplementary material.}
\label{tab:comparison_N}
\centering
\resizebox{\linewidth}{!}{%
\begin{tabular}{@{}cllccccc}
\toprule

\textbf{\# Layers}
& \textbf{Sampling Schedule}
& \textbf{Methods}
& \textbf{Speed.} $\uparrow$
& \makecell{\textbf{RGB}\\\textbf{L1}$^{\mathrm{avg}}$ $\downarrow$}
& \makecell{\textbf{Alpha soft}\\\textbf{IoU}$^{\mathrm{avg}}$ $\uparrow$}
& \textbf{PSNR} $\uparrow$
& \textbf{LPIPS} $\downarrow$ \\

\midrule


\multirow{4}{*}{2}
& \multirow{2}{*}{VDE-fast}
& VDE~\cite{Tan_2026_CVPR}
& \textbf{2.62}$\times$
& 0.1327 & 0.6225
& 25.73 & 0.1082 \\

& & \cellcolor{gray!15}\textbf{FastLayered}
& \cellcolor{gray!15}2.61$\times$
& \cellcolor{gray!15}\textbf{0.1271}
& \cellcolor{gray!15}\textbf{0.6274}
& \cellcolor{gray!15}\textbf{26.65}
& \cellcolor{gray!15}\textbf{0.1037} \\

\cdashline{2-8}[0.8pt/1.5pt]

& \multirow{2}{*}{MeanCache ($\mathcal{B}=10$)}
& MeanCache~\cite{gao2026meancache}
& \textbf{4.94}$\times$
& 0.1321 & 0.6206
& 26.01 & 0.1171 \\

& & \cellcolor{gray!15}\textbf{FastLayered}
& \cellcolor{gray!15}4.90$\times$
& \cellcolor{gray!15}\textbf{0.1281}
& \cellcolor{gray!15}\textbf{0.6253}
& \cellcolor{gray!15}\textbf{26.58}
& \cellcolor{gray!15}\textbf{0.1043} \\

\midrule


\multirow{4}{*}{3}
& \multirow{2}{*}{VDE-fast}
& VDE~\cite{Tan_2026_CVPR}
& \textbf{2.62}$\times$
& 0.1458 & 0.6098
& 25.16 & 0.1141 \\

& & \cellcolor{gray!15}\textbf{FastLayered}
& \cellcolor{gray!15}2.61$\times$
& \cellcolor{gray!15}\textbf{0.1403}
& \cellcolor{gray!15}\textbf{0.6182}
& \cellcolor{gray!15}\textbf{26.03}
& \cellcolor{gray!15}\textbf{0.1083} \\

\cdashline{2-8}[0.8pt/1.5pt]

& \multirow{2}{*}{MeanCache ($\mathcal{B}=10$)}
& MeanCache~\cite{gao2026meancache}
& \textbf{4.95}$\times$
& 0.1452 & 0.6092
& 25.55 & 0.1204 \\

& & \cellcolor{gray!15}\textbf{FastLayered}
& \cellcolor{gray!15}4.90$\times$
& \cellcolor{gray!15}\textbf{0.1408}
& \cellcolor{gray!15}\textbf{0.6189}
& \cellcolor{gray!15}\textbf{26.02}
& \cellcolor{gray!15}\textbf{0.1089} \\

\midrule


\multirow{4}{*}{4}
& \multirow{2}{*}{VDE-fast}
& VDE~\cite{Tan_2026_CVPR}
& \textbf{2.62}$\times$
& 0.1682 & 0.5721
& 24.73 & 0.1192 \\

& & \cellcolor{gray!15}\textbf{FastLayered}
& \cellcolor{gray!15}2.60$\times$
& \cellcolor{gray!15}\textbf{0.1626}
& \cellcolor{gray!15}\textbf{0.5815}
& \cellcolor{gray!15}\textbf{25.61}
& \cellcolor{gray!15}\textbf{0.1129} \\

\cdashline{2-8}[0.8pt/1.5pt]

& \multirow{2}{*}{MeanCache ($\mathcal{B}=10$)}
& MeanCache~\cite{gao2026meancache}
& \textbf{4.95}$\times$
& 0.1675 & 0.5727
& 25.03 & 0.1277 \\

& & \cellcolor{gray!15}\textbf{FastLayered}
& \cellcolor{gray!15}4.91$\times$
& \cellcolor{gray!15}\textbf{0.1629}
& \cellcolor{gray!15}\textbf{0.5809}
& \cellcolor{gray!15}\textbf{25.45}
& \cellcolor{gray!15}\textbf{0.1150} \\

\bottomrule
\end{tabular}%
}
\end{table}

\subsection{Quantitative Results}

\noindent\textbf{Main Results.}
Table~\ref{tab:quantitative_comparison} compares inference efficiency, layer decomposition accuracy, and global reconstruction fidelity on the Crello and Figma datasets. FastLayered achieves a \textbf{5.44$\times$} inference speedup on both datasets, with latencies of 23.66s and 23.62s, respectively. On Crello, FastLayered maintains layer decomposition quality comparable to $T=50$ and improves global reconstruction fidelity. The PSNR increases from \textbf{27.57 dB} to \textbf{27.70 dB}, and the LPIPS decreases from \textbf{0.0712} to \textbf{0.0707}. These results show that FastLayered achieves competitive layer decomposition and global reconstruction quality while substantially accelerating inference.

\begin{table}[t]
    \centering
    \caption{Ablation study of velocity prediction components on Crello under the MeanCache ($\mathcal{B}=10$) sampling schedule. ``+'' indicates that the corresponding component is added to the velocity reuse baseline.}
    \label{tab:velocity_ablation}
    \resizebox{\linewidth}{!}{
    \begin{tabular}{lcccc}
        \toprule
        \textbf{Methods} 
        & \makecell{\textbf{RGB}\\\textbf{L1}$^{\mathrm{avg}}$ $\downarrow$}
        & \makecell{\textbf{Alpha soft}\\\textbf{IoU}$^{\mathrm{avg}}$ $\uparrow$}
        & \textbf{PSNR} $\uparrow$ 
        & \textbf{LPIPS} $\downarrow$ \\
        \midrule
        Vanilla ($T=50$)~\cite{yin2026qwen}  
        & 0.1298 & 0.7789 & 27.57 & 0.0712 \\
        Velocity reuse 
        & 0.1301 & 0.7784 & 27.48 & 0.0729 \\
        + Composite Velocity Prediction 
        & 0.1299 & 0.7792 & 27.60 & 0.0717 \\
        + Output Layer Velocity direction prediction 
        & 0.1297 & 0.7803 & 27.54 & 0.0720 \\
        + Output Layer Velocity magnitude prediction 
        & 0.1295 & 0.7868 & 27.69 & 0.0710 \\
        + CGVC 
        & \textbf{0.1294} & \textbf{0.7873} & \textbf{27.73} & \textbf{0.0703} \\
        \bottomrule
    \end{tabular}
    }
\end{table}

\noindent\textbf{Comparison under the Same Sampling Schedules.}
To further compare velocity forecasting methods, we adopt the sampling schedules of VDE and MeanCache. As shown in Table~\ref{tab:quantitative_comparison_CGVC}, FastLayered achieves more accurate layer decomposition and higher-fidelity global reconstruction on Crello and Figma under all compared schedules. It also maintains comparable inference latency. Specifically, under the VDE-fast schedule on Crello, FastLayered \textbf{improves PSNR by 1.95 dB}. Under the MeanCache ($\mathcal{B}=10$) schedule on Figma, it \textbf{reduces LPIPS by 9.95\%}. These results demonstrate the effectiveness of FastLayered under different sampling schedules.

\begin{figure*}[t]
    \centering
    \includegraphics[width=\linewidth]{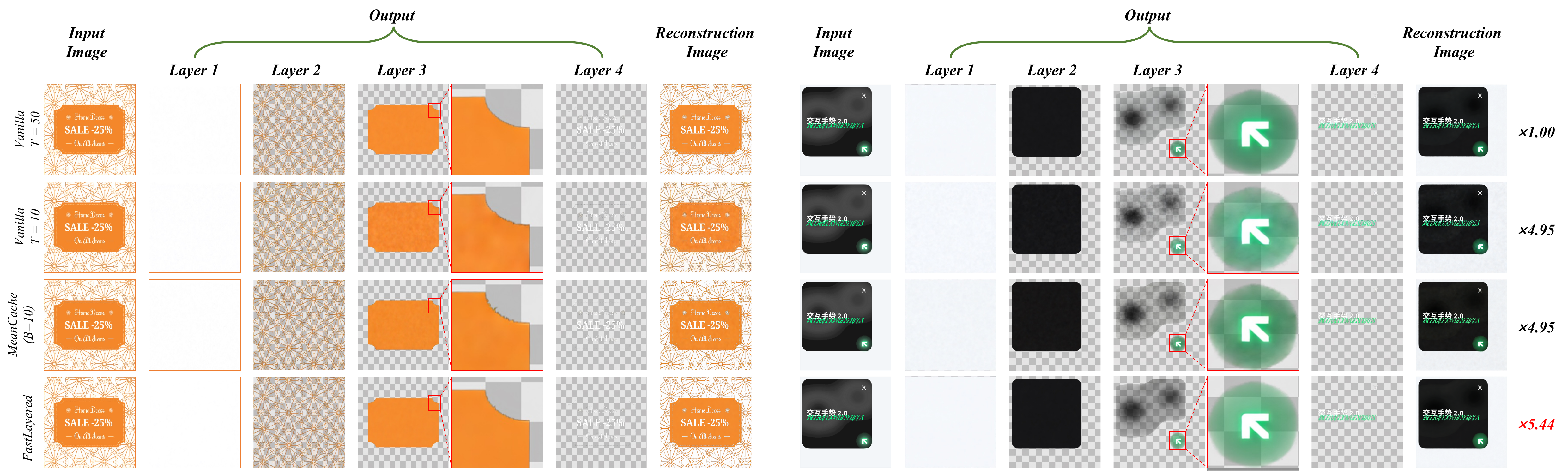}
    \caption{\textbf{Visual comparison of layer decomposition across different methods.} The examples on the left and right are from Crello and Figma, respectively. We show the input image, four output layers, and the reconstructed image for each method. The red boxes show enlarged regions of the output layers. The values on the right indicate inference speedups relative to vanilla inference with $T=50$.}
    \label{fig:layer_visualization}
\end{figure*}

\begin{figure*}[t]
    \centering
    \includegraphics[width=\linewidth]{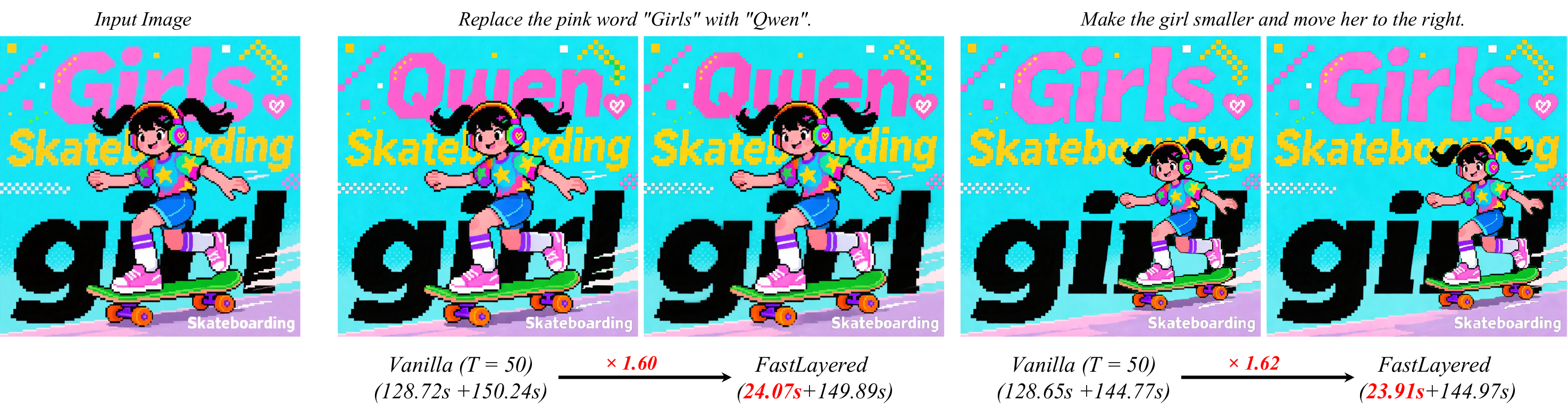}
    \caption{\textbf{Qualitative comparison of accelerated image editing.} We compare text replacement and character adjustment results using Vanilla and FastLayered. The values in parentheses indicate layer generation and editing times, respectively. The arrows show the overall speedups relative to Vanilla. More visual results are provided in the supplementary material.}
    \label{fig:edit_visualization}
\end{figure*}

\noindent\textbf{Different Numbers of Output Layers.}
Table~\ref{tab:comparison_N} compares the methods with different numbers of output layers. Under the same sampling schedules, FastLayered achieves better layer decomposition and global reconstruction quality with 2, 3, and 4 output layers. Specifically, under the MeanCache schedule with three output layers, FastLayered reduces the average RGB L1 from \textbf{0.1452} to \textbf{0.1408} and increases the PSNR from \textbf{25.55 dB} to \textbf{26.02 dB}.

\subsection{Ablation Studies}

We conduct ablation studies on Crello to analyze the contributions of the velocity prediction components. As shown in Table~\ref{tab:velocity_ablation}, each component improves layer decomposition and global reconstruction quality over the velocity reuse baseline. Specifically, historical velocity fusion increases the average Alpha soft IoU from \textbf{0.7784} to \textbf{0.7803}. Composite-guided magnitude calibration increases it to \textbf{0.7868} and reduces LPIPS from \textbf{0.0729} to \textbf{0.0710}. Notably, CGVC not only outperforms velocity reuse but also achieves competitive layer decomposition accuracy and global reconstruction fidelity compared with vanilla inference at $T=50$.

\subsection{Qualitative Results}

\noindent\textbf{Layer Decomposition Visualization.}
As shown in Fig.~\ref{fig:layer_visualization}, we compare the layer decomposition and reconstruction results of different methods. In the Crello example, the orange shapes generated by $T=10$ and MeanCache have locally uneven edges. FastLayered effectively alleviates this issue. In the Figma example, the layers generated by $T=10$ and MeanCache show uneven transparency or reduced local quality in the green icon and surrounding semi-transparent regions. FastLayered better preserves the visual quality of these regions. Overall, FastLayered better preserves layer details and achieves good global reconstruction quality while accelerating inference. More visual results are provided in the supplementary material.

\noindent\textbf{Image Editing Application.} As shown in Fig.~\ref{fig:edit_visualization}, the layers generated by FastLayered support text replacement and character adjustment, with editing results comparable to Vanilla. By accelerating layer generation, FastLayered achieves an overall speedup of \textbf{at least 1.60}$\times$.

\section{Conclusion}

In this paper, we propose FastLayered, a training-free acceleration method for image-to-layer generation. FastLayered is based on Composite-Guided Velocity Calibration (CGVC). CGVC uses the input-image latent as an approximate endpoint reference to predict the composite velocity. For each output layer, it combines historical velocities to predict the velocity direction and uses the relative magnitude change of the composite velocity to calibrate the velocity magnitude. Experiments on Crello and Figma show that FastLayered achieves a $5.44\times$ inference speedup. Under the same sampling schedules, it achieves better layer decomposition and global reconstruction quality than the compared velocity forecasting methods.

\subsection*{LLM Usage Statement} 
ChatGPT (OpenAI) was used to assist with editing and refining author-written manuscript text, including language, organization, formatting, and the presentation of related work. These uses supported manuscript preparation and revision, with the authors responsible for deciding whether and how to incorporate the suggestions. The authors retain full responsibility for all submitted content, including the methodology, experimental results, interpretations, conclusions, and the accuracy and appropriateness of the text and references.

{
    \small
    \bibliographystyle{ieeenat_fullname}
    \bibliography{main}
}


\end{document}